\documentclass[review,jog]{igs}

\usepackage[utf8]{inputenc}

\usepackage{mathabx}
\usepackage{graphicx}
\usepackage{siunitx}
\usepackage{tabularx}
\usepackage{makecell}
\usepackage{authblk}
\usepackage{multirow}
\usepackage{xcolor}
\usepackage{soul}

\jourvolume{}
\jourissue{}
\jourpubyear{2024}

\begin{document}

\title{Graph convolutional network as a fast statistical emulator for numerical ice sheet modeling}

\author[Koo et al.]{Younghyun Koo,$^{1,2}$
  Maryam Rahnemoonfar$^{1,2}$\protect\thanks{Present address:
  113 Research Drive, Lehigh University Mountaintop Campus, Bethlehem, PA 18015}}

\affiliation{%
$^1$Department of Computer Science and Engineering, Lehigh University, Bethlehem, PA, USA\\
$^2$Department of Civil and Environmental Engineering, Lehigh University, Bethlehem, PA, USA
  Correspondence: Maryam Rahnemoonfar
  \email{maryam@lehigh.edu}}

\begin{frontmatter}

\maketitle

\begin{abstract}
The Ice-sheet and Sea-level System Model (ISSM) provides numerical solutions for ice sheet dynamics using finite element and fine mesh adaption. However, considering ISSM is compatible only with central processing units (CPUs), it has limitations in economizing computational time to explore the linkage between climate forcings and ice dynamics. Although several deep learning emulators using graphic processing units (GPUs) have been proposed to accelerate ice sheet modeling, most of them rely on convolutional neural networks (CNNs) designed for regular grids. Since they are not appropriate for the irregular meshes of ISSM, we use a graph convolutional network (GCN) to replicate the adapted mesh structures of the ISSM. When applied to transient simulations of the Pine Island Glacier (PIG), Antarctica, the GCN successfully reproduces ice thickness and velocity with a correlation coefficient of approximately 0.997, outperforming non-graph models, including fully convolutional network (FCN) and multi-layer perceptron (MLP). Compared to the fixed-resolution approach of the FCN, the flexible-resolution structure of the GCN accurately captures detailed ice dynamics in fast-ice regions. By leveraging 60-100 times faster computational time of the GPU-based GCN emulator, we efficiently examine the impacts of basal melting rates on the ice sheet dynamics in the PIG.



\end{abstract}

\end{frontmatter}


\section{ Introduction}

\begin{figure}
    \centering
    \includegraphics[width=0.9\linewidth]{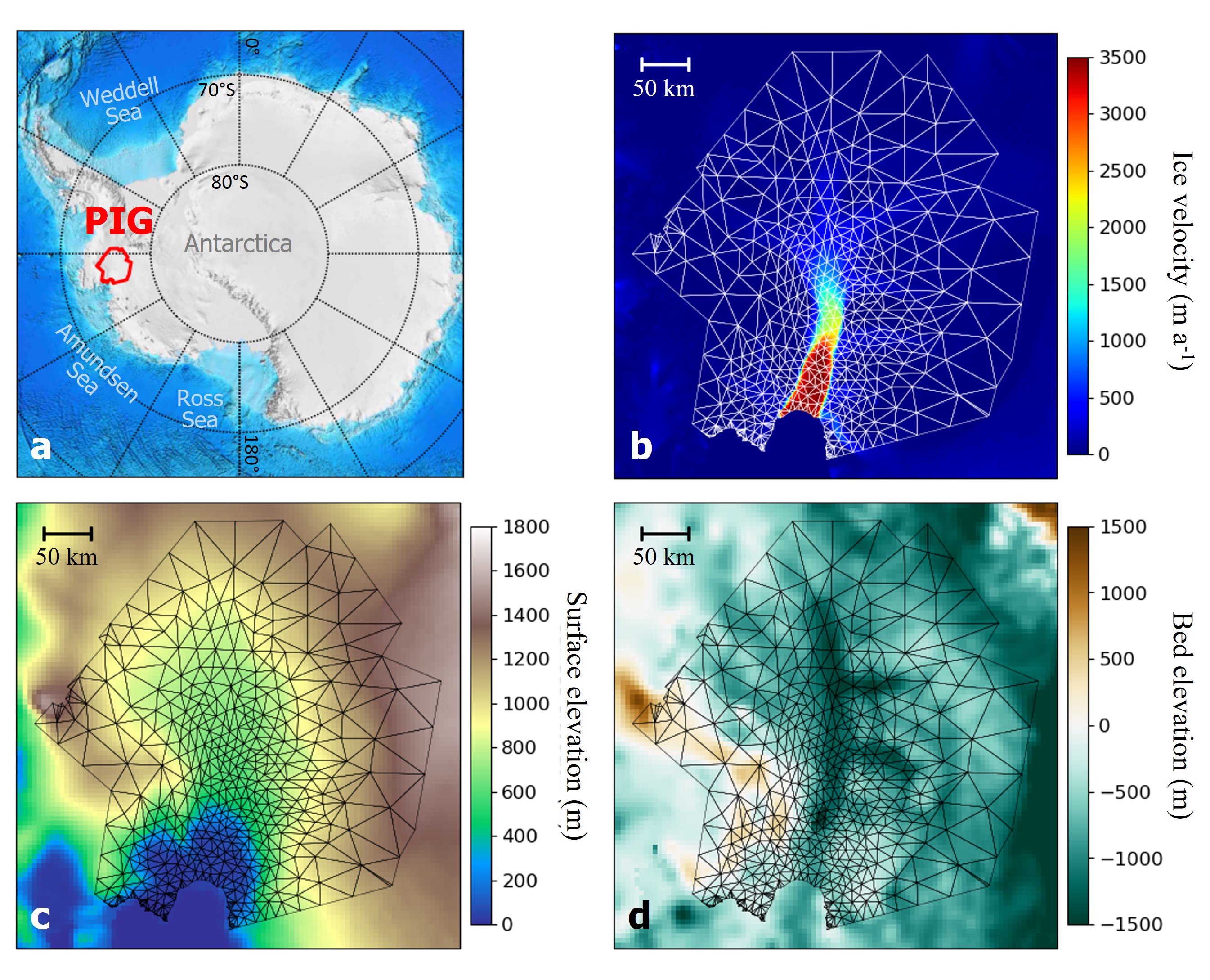} 
    \caption{(a) Location of Pine Island Glacier (PIG) in the Antarctic indicated by a red poylgon. Dashed lines are 10-degree-apart latitudes and 30-degree-apart longitudes. (b) Initial ice velocity, (b) surface elevation, and (c) ice thickness of the PIG. The meshes in (b), (c), and (d) are initialized with 20 km mesh size. The extent of meshes in (b), (c), and (d) correspond to the red polygon in (a).}
    \label{PIG}
\end{figure}

As the global climate has been warming, ice sheets in Greenland and the Antarctic have lost more than 7500 Gt of ice from 1992 to 2020, contributing to approximately 21 mm of global sea-level rise \citep{Otosaka2023}. The rate of ice loss is now up to five times higher than in the early 1990s in Greenland and 25 \% higher in the Antarctic. In Antarctica, the Pine Island Glacier (PIG) (Fig. \ref{PIG}a) has experienced the most rapid mass loss and acceleration in ice velocity, which were primarily induced by melt-driven thinning near the grounding line and calving events \citep{Joughin2021_2, Joughin2021, jacobs2011}. Consequently, the PIG accounts for more than 20 \% of Antarctica’s contribution to global sea-level rise \citep{Rignot2019}. Due to the great contribution of the PIG to global sea-level rise, accurate modeling of ice thickness and velocity of the PIG has been a paramount concern for the science community \citep{Laour2012_PIG, Seroussi2014_PIG, Rosier2021, vieli2003, Gladstone2012}. 


For the last few decades, scientists have proposed several physical models to explain the thermomechanical flows of ice sheets. In those modeling studies, ice is the viscous non-Newtonian fluid that follows the Stokes equation \citep{Glen1955}. On the basis of the Stokes equation, the dynamic mechanisms of large ice sheets can be described by ``Full-Stokes'' (FS) equations consisting of partial differential equations (PDEs) with four unknowns: the 3 components of the ice velocity, $(u, v, w)$, and the pressure, $p$. However, since solving FS equations is computationally expensive and impractical at a continental scale and a high resolution, several simplified approximations of FS have been proposed. For example, the Shallow Ice Approximation (SIA) \citep{Hutter1983} assumes that ice sheet dynamics is mostly driven by basal sheer stress, balancing the basal shear stress and gravitational driving stress of grounded ice. However, its simplified mechanical assumptions limit its applicability to non-ice-streaming regions and valley glaciers where ice flow is dominated by vertical shearing. For ice streams or floating ice shelves, the Shallow Shelf Approximation (SSA) \citep{Morland1987_SSA, MacAyeal1989} provides an alternative 2D solution by assuming that horizontal velocity is depth-independent and vertical shear stresses are negligible. However, SSA cannot describe the ice dynamics in areas where vertical variations in speed are considerable, such as grounding lines, ice stream margins, or complex ice flow near an ice divide. Besides the 1D solutions of SIA and 2D solutions of SSA, the Blatter-Pattyn approximation (BP) \citep{Blatter1995, Pattyn1996} provides valid and efficient 3D solutions in the majority of an ice sheet, both longitudinal stresses of fast-flowing ice streams and vertical shear stresses of slow ice.

The Ice-sheet and Sea-level System Model (ISSM) is a thermomechanical numerical model that provides solutions for these four different ice flow models using a finite element approach \citep{Larour2012}. The unique characteristics of the ISSM model can be summarized as follows: (1) finite element methods, (2) fine mesh adaptation, and (3) parallel technologies. First, by using unstructured meshes, the ISSM provides efficient ice flow solutions with high resolutions in areas where ice flow dynamics is critical. Second, adaptive mesh refinement (AMR) allows ISSM to allocate its computational resources to the fine-resolution area of fast ice and coarse-resolution areas of stagnant ice. Finally, state-of-the-art parallel technologies reduce the running time of the ISSM model dramatically when implemented in computer clusters.

Despite these advantages of ISSM in reducing computational time via parallel technologies, implementing the ISSM model is only available through multi-core central processing units (CPUs) because solving PDEs of ice dynamics requires serial processing of CPU. The computational demand of CPU-based numerical modeling makes it time-consuming and inefficient to explore the impacts of climate forcings on ice sheet dynamics. Hence, in order to predict the ice sheet dynamics under various climate forcings (e.g., temperature, CO2 concentration, basal melting rate), statistical emulators have been used to approximate the mapping between climate forcings and ice sheet behaviors \citep{Katwyk2023_emulator, Edwards2021_emulator, Berdahl2021_emulator}. Recently, leveraging the capability of graphic processing units (GPUs) in parallel processing, deep learning techniques have emerged as an attractive and efficient tool for statistical emulators. GPUs divide a given task into a number of small tasks and process them in parallel, which allows a considerable speed-up compared to serial processing by CPU. Although GPUs cannot be used to directly solve the PDEs of ice flow, the parallel processing ability of GPUs for matrix multiplication allows deep learning models to learn the statistical features of numerical simulation results with much faster computational time. Thus, once trained with numerical simulation data, GPU-based deep learning models can act as emulators that replicate the behavior of numerical models and accelerate the computational time of ice sheet modeling. 

\begin{figure}
    \centering
    \includegraphics[width=0.9\linewidth]{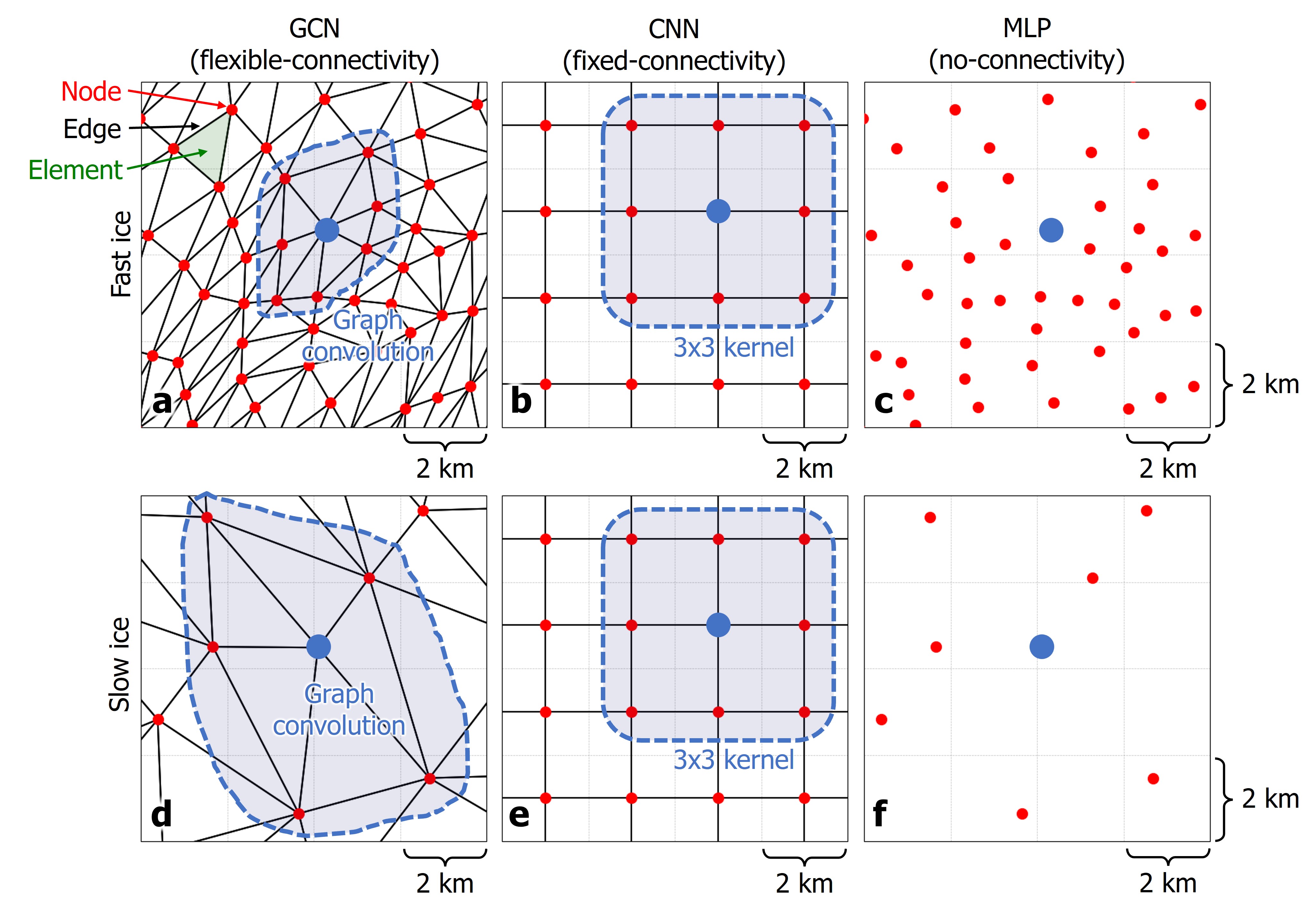}
    \caption{Structures of data and node connectivity for different deep learning architectures: graph convolutional network (GCN), convolutional neural network (CNN), and multi-layer perceptron (MLP) for fast and slow ice conditions. The graph structure of GCN is converted from the finite element (i.e., unstructured meshes) of ISSM: the fast-ice area has a fine mesh resolution (a), and the slow-ice area has a coarse mesh resolution (d), and the nodes of the element are connected as edges. The graph convolution of a node is determined by the neighboring nodes. The regular grid structure of CNN has the same resolution (2 km) for all locations regardless of ice velocity (b and e), and the convolutional kernel size is fixed to 3$\times$3 for all locations. The node of MLP is the same as the node of ISSM and GCN, but the connectivity between nodes is not used; only the features of a node are used to predict the ice condition at that node (c and f).}
    \label{graph}
\end{figure}

When using deep learning techniques as statistical emulators for numerical ice sheet models, it is essential to select an appropriate model architecture to represent the geospatial features of ice sheets accurately. In this aspect, convolutional neural networks (CNNs) have been employed as a typical architecture to capture the spatial variations in topographical features, which play a key role in determining ice sheet dynamics \citep{Jouvet2022, jouvet2023_PIML, jouvet2023_inversion}. However, although traditional CNN architectures can successfully recognize spatial patterns of Euclidean or grid-like structures (e.g., images) by using fixed-size trainable localized filters, they cannot be used for non-Euclidean or irregular structures where the connections to neighbors are not fixed \citep{zhang2019}. Since ISSM uses unstructured meshes, using CNN as an emulator for ISSM can introduce two problems: (1) the CNN grid with fixed resolution can lose dynamical details in fast ice areas; (2) the CNN grid requires unnecessary computational demands in slow ice areas. Instead, for such non-Euclidean or irregular data structures, including molecules, point clouds, social networks, and natural language processing, graph neural networks (GNNs) have been proposed and broadly utilized \citep{zhang2019}. In GNNs, data structures are depicted as graphs consisting of data points (i.e., \emph{nodes}) connected by lines (i.e., \emph{edges}). GNNs find statistical patterns or make predictions in graphs using pairwise message passing between nodes, such that individual nodes iteratively update their representations by exchanging information with their neighboring nodes. Various GNN architectures have been proposed with various message passing procedures by different applications and graph characteristics \citep{zhang2019}. Among these, graph convolutional network (GCN), the most typical and simple GNN architecture, can replace traditional CNNs in non-regular graph structures because it applies convolutional operators directly on graphs (Fig. \ref{graph}).

The objective of this study is to develop a computationally efficient GCN emulator for ISSM in order to infer the contributions of thermal melting to the ice dynamics of the PIG. We substitute the ISSM meshes with graph structures and let our GCN imitate the ability of the ISSM to predict ice thickness and flow. The main contributions of this study are the following:
\begin{itemize}
    \item We develop a GCN as a statistical emulator to reproduce the ice thickness and velocity simulated from the ISSM numerical ice sheet model.
    \item We conduct extensive experiments for the PIG to evaluate the fidelity and performance of the GCN model, and compare the GCN with other non-graph machine learning models.
    \item Using the fast GCN emulator, we examine the impacts of basal melting rate on the ice volume and velocity in the PIG.
\end{itemize}

 
The remainder of the paper is organized as follows. Section \ref{method} describes the details of how to train and evaluate a GCN and other baseline deep learning models with ISSM simulation data. Next, in section \ref{results}, we evaluate the fidelity and computational performance of the GCN emulator by comparing it with CPU-based ISSM simulations and other baseline models. Section \ref{results} also demonstrates how this GCN emulator can be used to investigate the impacts of basal melting rate on the ice sheet behavior. In section \ref{discussion}, we evaluate the upstream training costs of the deep learning emulators and provide a comprehensive discussion about the advantages and challenges of the GCN.



\section{ Methods}\label{method}


We train and test a GCN emulator with the numerical ice sheet simulations of ISSM. We compare this GCN emulator with two non-graph deep learning architectures: CNN and multi-layer perceptron (MLP).


\subsection{ Preparation of training data from numerical simulations}

In order to train deep learning emulators, we collect the numerical simulation results in the PIG from the ISSM transient simulations. We simulate the 20-year evolution of ice thickness and ice velocity in the PIG by adapting the ISSM-based sensitivity experiments conducted by \cite{Seroussi2014_PIG} and \cite{Laour2012_PIG}. Since the PIG has a significant portion of floating ice, we use the SSA \citep{Morland1987_SSA, MacAyeal1989} to explain ice flow. The SSA can be expressed by the following equations:

\begin{equation}\label{SSA1}
    \frac{\partial}{\partial x}\left(4H\mu\frac{\partial u}{\partial x}+2H\mu\frac{\partial v}{\partial y}\right)+\frac{\partial}{\partial y}\left(H\mu\frac{\partial u}{\partial y}+H\mu\frac{\partial v}{\partial x}\right) = \rho g H \frac{\partial s}{\partial x}
\end{equation}
\begin{equation}\label{SSA2}
    \frac{\partial}{\partial y}\left(4H\mu\frac{\partial v}{\partial y}+2H\mu\frac{\partial u}{\partial x}\right)+\frac{\partial}{\partial x}\left(H\mu\frac{\partial u}{\partial y}+H\mu\frac{\partial v}{\partial x}\right) = \rho g H \frac{\partial s}{\partial y}
\end{equation}
where $(u, v)$ are the $x$ and $y$ components of the ice velocity vector in the Cartesian coordinate system, $H$ is the local ice thickness, $\mu$ is the effective ice viscosity,  $\rho$ is ice density, and $g$ is the acceleration due to gravity.

The ISSM transient simulations require some ice sheet observation data (e.g., ice velocity, thickness, and bed elevation) and climatological data (e.g., temperature, surface mass balance). We collect ice velocity data from the NASA Making Earth System Data Records for Use in Research Environments (MEaSUREs) \citep{MEASURE_velocity_SAR} (Fig. \ref{PIG}b). This ice velocity map is derived from multiple SAR images from 2007 to 2009 with 1 km spatial resolution. The surface elevation and bed topography are collected from the 1 km Antarctic digital elevation model (DEM) data in 2008 \citep{Bamber2009_DEM} (Fig. \ref{PIG}c) and the Amundsen Sea bathymetric map data in 2007 with 250 m grids \citep{Nitsche2007_bedrock} (Fig. \ref{PIG}d). Finally, we collect three climatological datasets: Antarctic surface temperature records of infrared satellite data (1 km resolution) and weather stations from 1979 to 1998 \citep{Comiso2000_temperature}, 5-km interpolated Antarctic surface mass balance (SMB) map in 1986-1998 \citep{Vaughan1999_SMB}, and 5-km interpolated Antarctic geothermal heat flux in 2005 \citep{Maule2005_heatflux}.

The unstructured meshes of ISSM are generated and adapted by the bi-dimensional anisotropic mesh generator (BAMG) \citep{hecht_bamg}. Once a triangular mesh is set with an initial mesh size ($M_0$), the BAMG algorithm refines the mesh by splitting the triangle edges and inserting new vertices in the mesh until the desired resolution is reached. The desired resolution is determined by criteria based on the element distance to the grounding line and ZZ error estimator for deviatoric stress tensor and ice thickness \citep{ISSM_mesh_2019}. This AMR procedure can improve the accuracy of numerical simulations and reduce computational cost compared to the uniform meshes without AMR \citep{ISSM_mesh_2019}. To examine how the results change with mesh resolution, we implement the ISSM simulations on three different mesh sizes: $M_0$=2 km, $M_0$=5 km, and $M_0$=10 km. For different mesh size experiments, meshes are adjusted by ice velocity after the initialization: the area with fast ice has a relatively finer mesh resolution, and slow ice has a coarser mesh resolution \citep{ISSM_mesh_2019, Larour2012}. The 2 km, 5 km, and 10 km mesh initialization generates the final mesh grid with 12,459, 5,499, and 3,511 elements, respectively, corresponding to 6,384, 2,852, and 1,833 nodes (Table \ref{table_nodes_elements}).

\begin{table}
\caption{The number of nodes, edges, and elements for three $M_0$ settings}
\begin{tabular}[t]{c|c|c|c}
\hline
$M_0$ & 2 km & 5 km & 10 km \\
\hline
Nodes & 6,384 & 2,852 & 1,833 \\
Edges & 37,684 & 16,700 & 10,686 \\
Elements & 12,459 & 5,499 & 3,511 \\
\hline
\end{tabular}
\label{table_nodes_elements}
\end{table}%


In addition, considering basal melting is the main driver of ice mass loss in the PIG \citep{Joughin2021_2, jacobs2011}, we collect the ISSM simulations from different basal melting rate scenarios to statistically examine how different melting rates change the dynamic behavior of ice sheet. We implement the ISSM simulations for 36 different annual basal melting rates ranging from 0 to 70 m $\text{a}^{-1}$ for every 2 m $\text{a}^{-1}$. Transient simulations are run forward for 20 years with time steps of one month. Consequently, we execute 20-year (240-month) transient simulations 108 times: 3 different mesh sizes and 36 different melting rates.

The ISSM simulation produces the ice velocity and thickness predictions for individual nodes that consist of adjusted triangular meshes (Fig. \ref{graph}). In order to use this data as the input and output training data for GCN, CNN, and MLP, we convert the raw mesh of ISSM into a certain data structure that corresponds to each deep learning architecture. For the GCN architecture, we convert the raw finite elements into graph nodes and edges. In the ISSM meshes, each element consists of three nodes; we connect these nodes with edges. Since we use the nodes and elements of the raw ISSM mesh, the resolution of this graph is exactly the same as the ISSM simulation. On the other hand, since the CNN requires regular grid data, we interpolate the raw ISSM mesh into a 2 km regular grid using bilinear interpolation. Since the 2 km resolution is applied to all locations, the resolution of this regular grid can be more coarse than the raw ISSM mesh in the fast-ice region (Fig. \ref{graph}a and b) but finer in the slow-ice region (Fig. \ref{graph}d and e). The MLP uses the same nodes as the GCN, but it does not incorporate the connectivity information between the nodes (Fig. \ref{graph}c and f).



\subsection{ Graph Convolutional Network (GCN)}

We experiment with the GCN architecture proposed by \cite{Kipf2016_GCN}. In this multi-layer GCN architecture, let the undirected graph $\mathcal{G} = (\mathcal{V}, \mathcal{E})$ consist of $N$ nodes $v_i \in \mathcal{V}$ $(1\leq i \leq N)$ and edges $(v_i, v_j) \in \mathcal{E}$ $(1\leq i,j \leq N)$. The connectivity between nodes $v_i$ and $v_j$ can be represented by an adjacency matrix $A \in \mathbb{R}^{N \times N}$. When the node features in the $l$th layer are propagated to the ($l+1$)th layer, the node features are updated using the following layer-wise propagation rule:

\begin{equation}
\label{eq_gcn_propogation}
H^{(l+1)}=\sigma(\Tilde{D}^{-\frac{1}{2}}\Tilde{A}\Tilde{D}^{-\frac{1}{2}}H^{(l)}W^{(l)})
\end{equation}
where $\Tilde{A}=A+I_N$ is the adjacency matrix of the undirected graph $\mathcal{G}$ with added self-connections. $I_N$ is the identity matrix, $\Tilde{D}_{ii}=\sum_{j}\Tilde{A}_{ij}$, $W$ is a layer-specific trainable weight matrix, and $\sigma(\cdot)$ is an activation function. $H^{(l)} \in \mathbb{R}^{N \times D}$ is the matrix of activations in the $l$ th layer, and $H^{(0)}$ is the input of the neural network. This propagation rule is inspired by the localized first-order approximation of spectral graph convolutions on graph-structured data \citep{Kipf2016_GCN}.

Let the $l$th graph convolutional layer receive a set of node features $H^{(l)} = \{h_1^{(l)}, h_2^{(l)}, ..., h_N^{(l)}\}$, $h_i^{(l)} \in \mathbb{R}^{F_{l}}$ as the input and produce a new set of node features, $H^{(l+1)} = \{h_1^{(l+1)}, h_2^{(l+1)}, ..., h_N^{(l+1)}\}$, $h_i^{(l+1)} \in \mathbb{R}^{F_{l+1}}$, for the ($l+1$)th layer. $F_{l}$ and $F_{l+1}$ are the number of features in each node at the $l$th layer and ($l+1$)th layer, respectively. Then, the layer-wise propagation rule of Eq. \ref{eq_gcn_propogation} can be expressed as follows:

\begin{equation}
\label{eq_gcn}
h_i^{(l+1)}=\sigma(\sum_{j\in\mathcal{N}(i)}\frac{1}{c_{ij}}W^{(l)}h_j^{(l)})
\end{equation}
where $\mathcal{N}(i)$ is the set of neighbors of $i$th node, $c_{ij}$ is an appropriately chosen normalization constant for the edge $(v_i, v_j)$ defined as the product of the square root of node degrees (i.e., $c_{ij}=\sqrt{|\mathcal{N}(j|)}\sqrt{|\mathcal{N}(i)|}$), and $W^{(l)} \in \mathbb{R}^{F_{l+1} \times F_{l}}$.

\begin{figure}
    \includegraphics[width=1.0\linewidth]{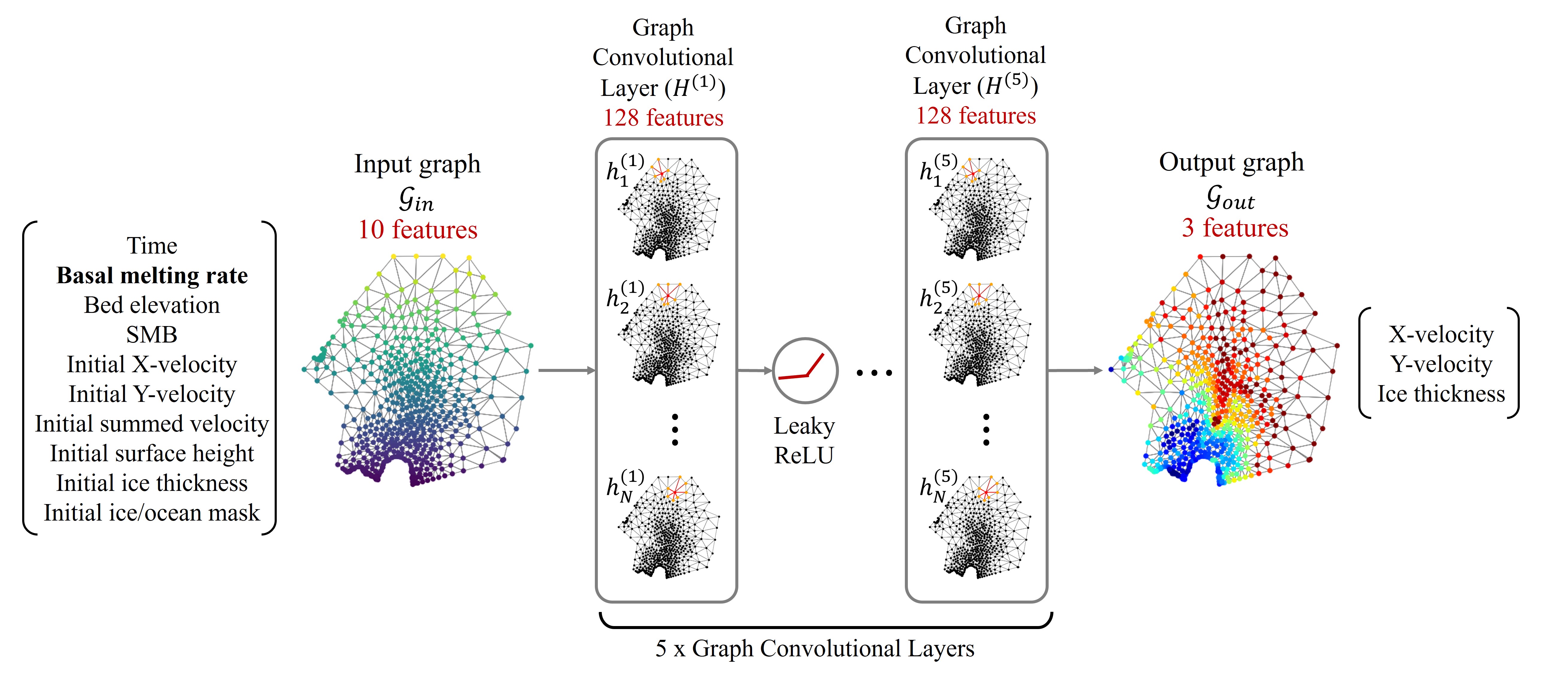} 
    \caption{Schematic illustration of the graph convolutional network (GCN) emulator.}
    \label{GNN}
\end{figure}

The graph structure $\mathcal{G}$ for the GCN is generated from unstructured meshes of ISSM; the nodes and edges of the meshes are taken as the nodes and edges of graph $\mathcal{G}$ (Fig. \ref{graph} and \ref{GNN}). To determine the optimal settings for the number of hidden layers and features of the GCN, trial-and-error experiments were conducted based on the mean square error (MSE). After we calculated MSE for 16 settings with 4 different hidden layers (1, 2, 5, and 10 layers) and 4 different numbers of features (32, 64, 128, and 256), we determined to use the combination of 5 hidden layers and 128 features, which produced the lowest MSE. The leaky Rectified Linear Units (leaky ReLU) function with 0.01 negative slope is chosen as the activation function (Fig. \ref{GNN}).


\subsection{ Baseline non-graph deep learning emulators}

The GCN, FCN, and MLP have significant differences in handling the spatial information of nodes. First, in the GCN, the nodes are iteratively updated through a series of graph convolutional layers by exchanging messages between the neighboring nodes connected by edges (Eq. \ref{eq_gcn}). As the most commonly used deep learning architecture for ice sheet emulators \citep{jouvet2023_inversion, jouvet2023_PIML, Jouvet2022}, the CNN can intrinsically embrace the neighboring information between nodes in regular grids through weights of the convolutional kernels, but the kernel size is fixed for all locations. On the other hand, as the most basic deep learning architecture \citep{popescu2009_mlp}, the MLP handles individual nodes independently and uses only the features of individual nodes to predict the output features; the connections between nodes are not embedded in the MLP architecture. In summary, the GCN uses the neighboring information between nodes through their node-edge connectivity regardless of their spatial distances (i.e., flexible connectivity); the CNN uses the fixed neighboring information for all nodes merely based on their distances (i.e., fixed connectivity); the MLP does not use any spatial neighboring information between nodes (i.e., no connectivity) (Fig. \ref{graph}). The detailed architectures of CNN and MLP are described in the following sections.


\subsubsection{ Convolutional Neural Network (CNN)}

The CNN in this study, particularly a fully convolutional network (FCN), has a similar architecture to \cite{Jouvet2022} but consists of one input layer (10 features), five hidden convolutional layers, and one output layer (3 features). The 2-D convolutional layers of this FCN have a kernel size of 3$\times$3 and a filter size of 128.  We conducted trial-and-error experiments to determine the optimal number of hidden layers and filters. We calculated MSE for 12 settings with 3 hidden layers (2, 5, and 10 layers) and 4 features per hidden layers (32, 64, 128, and 256), and 5 hidden layers and 128 features are determined as the best setting to minimize MSE and computational load of the FCN. Convolutional operations at each hidden layer multiply 3$\times$3 weights and pass to the next layer, and the leaky ReLU activation function of 0.01 negative slope is applied after each convolutional layer. Since the FCN takes regular grids as the input and output, the interpolated 2 km grid datasets are used as the input and output of the FCN.


\subsubsection{ Multi-layer perceptron (MLP)}

The MLP consists of one input layer (10 features), five hidden layers (128 features), and one output layer (3 features). When determining the hyperparameters of the MLP, we did not conduct trial-and-error experiments but used the same settings as the GCN. Maintaining an equal number of learning parameters for the GCN and MLP allows us to examine the pure effects of embedding adjacency information between nodes into the model training. All of the MLP layers are fully connected layers \citep{haykin1998_NN}: i.e., each node in hidden layers is updated with its own features without embedding the neighboring node features during the propagation process. The leaky ReLU activation function of 0.01 negative slope is applied after each hidden layer. 

\subsection{ Training and testing graph neural networks}

To train the GCN model, a total of 25,920 graph structures (240 months $\times$ 3 mesh sizes $\times$ 36 basal melting rates) are collected from the ISSM transient simulations. All nodes of the input graphs contain 10 input features (time, basal melting rate, bed elevation, surface mass balance (SMB), initial x-component velocity, initial y-component velocity, magnitude of initial velocity, initial surface height, initial ice thickness, and initial ice/ocean mask) and 3 output features (x-component ice velocity, y-component ice velocity, and ice thickness). We normalize the input and output feature values between [-1, 1] using the nominal maximum and minimum values that each variable can have. We divide the 25,920 graph structures into train, validation, and test datasets based on the melting rate values: melting rates of 0, 20, 40, and 60 m $\text{a}^{-1}$ are used for validation, melting rates of 10, 30, 50, and 70 m $\text{a}^{-1}$ are used for testing, and the remainders are for training. As a result, the number of train, validation, and test datasets is 20,160 (77.78 \%), 2,880 (11.11 \%), and 2,880 (11.11 \%), respectively. This data splitting approach allows us to assess how accurately our emulators can predict the ice sheet behaviors under out-of-training melting rate scenarios. We use the MSE as the loss function, and the model is optimized by Adam stochastic gradient descent algorithm \citep{kingma2017adam} with 500 epochs and 0.001 learning rate. All deep learning models are trained on the Python environment using the Deep Graph Library (DGL) \citep{wang2019_dgl} and PyTorch \citep{paszke2019_pytorch} modules. In measuring the computational time, we record the time to generate the final results of 20-year ice thickness and velocity for all 36 melting rates. Two computational resources of the same desktop (Lenovo Legion T5 26IOB6) are compared: a CPU (Intel(R) Core(TM) i7-11700F) and a GPU (NVIDIA GeForce RTX 3070).

\subsection{ Model evaluation}

We evaluate the ability of our emulators to reproduce ice velocity and ice thickness by comparing the predictions with the ISSM simulation results. We calculate two metrics for this assessment: root mean square error (RMSE) and correlation coefficient (R):

\begin{equation}
\text{RMSE}(\hat{y}, y)=\sqrt{\frac{1}{N}\displaystyle\sum_{i=1}^N(\hat{y}_i-y_i)^2}
\label{RMSE}
\end{equation}
\begin{equation}
\text{R}(\hat{y}, y)=\frac{\displaystyle\sum^N(\hat{y}_i-\bar{\hat{y}})(y_i-\bar{y})}{\sqrt{\displaystyle\sum^N(\hat{y}_i-\bar{\hat{y}})^2\sum^i(y_i-\bar{y})^2}}
\label{R}
\end{equation}
where $y$ is the reference value from ISSM simulations, $\hat{y}$ is the predicted value from emulators, and $N$ is the number of nodes. RMSE measures the difference between predictions and ISSM results; a value of 0 indicates the perfect fit between them. R measures the statistical correlation between predictions and ISSM results, ranging from -1 to 1; as a model better represents the spatial and temporal patterns of the ISSM results, the value is closer to 1. We note that we conduct additional interpolation to calculate the RMSE and R of the FCN model. Since the output of the FCN (2-km regular grid) does not exactly correspond to ISSM meshes, we interpolate the FCN outputs to the ISSM mesh using bilinear interpolation, and these interpolated values are compared to the true values of ISSM meshes.

\section{ Results}\label{results}

We evaluate the fidelity and computational efficiency of these deep learning emulators in predicting ice thickness and velocity. This evaluation is conducted and compared for three different initial mesh sizes ($M_0$) to investigate the sensitivity of results to mesh resolutions.


\subsection{ Ice thickness}


Table \ref{table1} shows the overall accuracy of ice thickness represented by RMSE and R. All deep learning emulators exhibit good agreements with the ISSM simulations with R greater than 0.999. On finer meshes of 2 km and 5 km $M_0$, GCN shows the best accuracy with around 12 m of RMSE. Even though the FCN shows the lowest RMSE on 10 km $M_0$, the difference between FCN and GCN is negligible. The MLP always shows the lowest accuracy among the three emulators on all mesh conditions. When comparing the accuracy for different mesh sizes, it is interesting that the RMSE of GCN and MLP remain consistent for different mesh resolutions: 12-14 m of RMSE for GCN and 24-25 m of RMSE for MLP. However, the RMSE of FCN varies from 14 to 21 m depending on the mesh resolutions; the FCN shows a lower fidelity on a finer mesh structure. This variation in accuracy with the FCN model could be attributed to the fixed regular grid structure of FCN. While the GCN and MLP directly take the unstructured meshes of ISSM (i.e., smaller mesh sizes for faster ice) as input and output, the FCN uses the 2 km regular grids interpolated from the ISSM meshes. Hence, this interpolated grid causes the FCN to lose topographical details and deteriorate the accuracy where the unstructured meshes are arranged more densely than the 2 km grid. Such a loss of spatial details can be critical in the precise delineation of the ice boundary where the ice conditions change drastically.


\begin{table}
\caption{Accuracy of ice thickness for three deep learning emulators with different mesh resolutions. All metrics are averaged for the 10, 30, 50, and 70 m $\text{a}^{-1}$ melting rates. The best accuracy for each mesh size is highlighted in bold.}
\begin{tabular}[t]{c|c|c|c|c|c|c}
\hline
\multirow{2}{*}{Model} &
\multicolumn{2}{c|}{$M_0$ = 2 km} &
\multicolumn{2}{c|}{$M_0$ = 5 km} &
\multicolumn{2}{c}{$M_0$ = 10 km} \\
\cline{2-7}
&RMSE (m) &R &RMSE (m) &R &RMSE (m) &R\\
\hline
GCN &\textbf{12.261} &\textbf{0.9998} &\textbf{12.678} &\textbf{0.9998} &14.267 &0.9997 \\
FCN &21.463 &0.9995 &17.120 &0.9997 &\textbf{14.248} &\textbf{0.9997}\\
MLP &24.125 &0.9992 &24.569 &0.9992 &25.337 &0.9991\\
\hline
\end{tabular}
\label{table1}
\end{table}%

\begin{figure}
    \centering
    \includegraphics[width=0.9\linewidth]{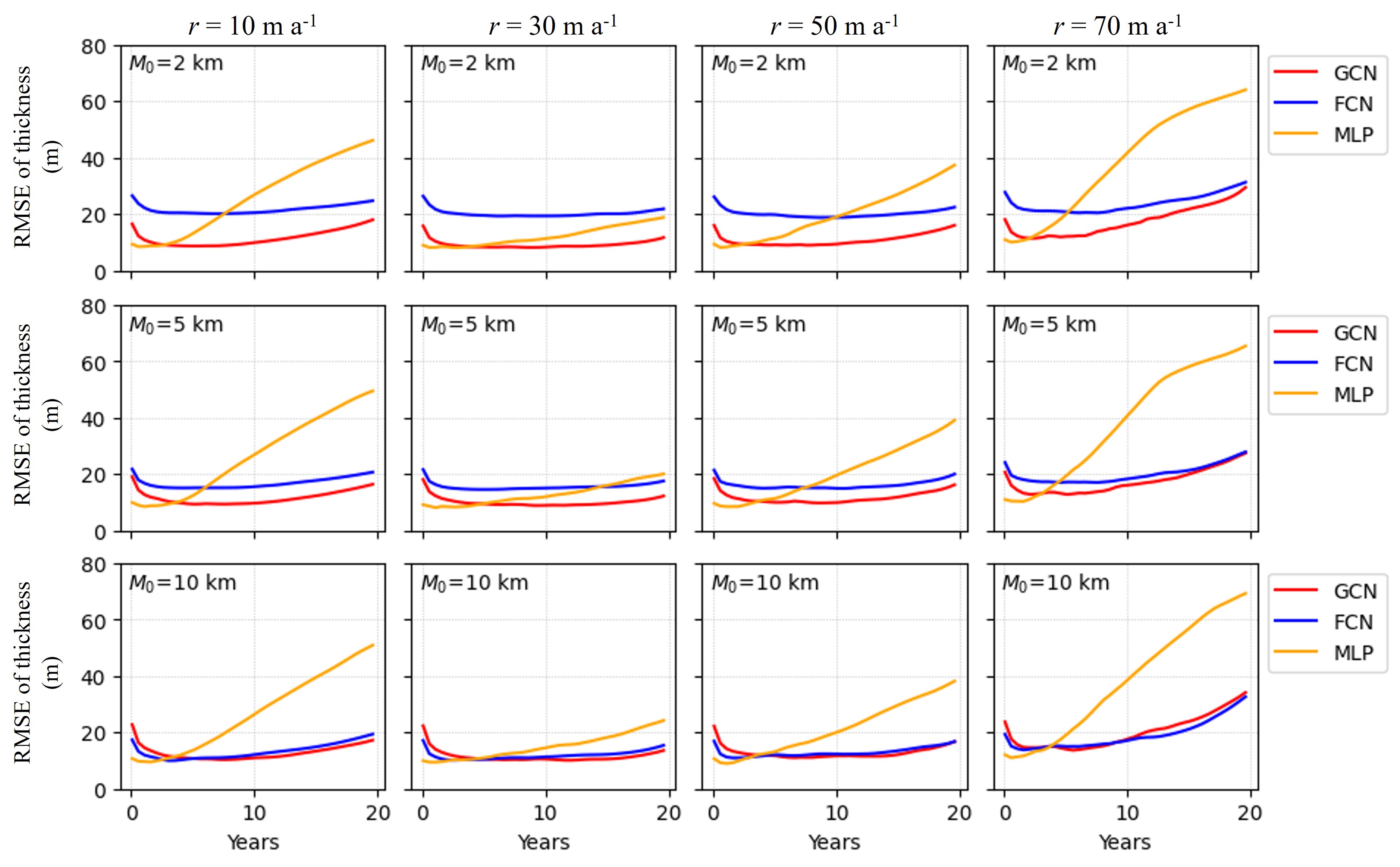} 
    \caption{RMSE of ice thickness by years for different basal melting rates ($r$) and initial mesh sizes ($M_0$).}
    \label{Thickness_year}
\end{figure}

Figure \ref{Thickness_year} shows the RMSE of ice thickness by years for four test melting rates ($r$ = 10, 30, 50, and 70 m $\text{a}^{-1}$) and three initial mesh sizes ($M_0$ = 2 km, 5 km, and 10 km). The GCN generally shows the lowest RMSE over 20-year transient simulation for most melting rates and mesh sizes. As shown in Figure \ref{Thickness_year}, the RMSEs of GCN and MLP do not vary significantly by different mesh sizes; however, the RMSE of FCN decreases with coarser mesh resolutions. In all melting rate scenarios, the RMSEs of GCN and FCN slightly decrease from the initial condition to around year 10 and then increase until year 20. On the contrary, the RMSE of MLP continuously increases from the initial condition to year 20. At the melting rate of 70 m $\text{a}^{-1}$, the RMSE of MLP reaches 60 m in year 20. The RMSE of ice thickness for a 70 m $\text{a}^{-1}$ melting rate experiment rises more rapidly by year compared to the other melting rates. Dramatic changes in ice flow under a higher melting rate scenario could make it difficult for deep learning emulators to replicate such a rapid change.

\begin{figure}
    \centering
    \includegraphics[width=1.0\linewidth]{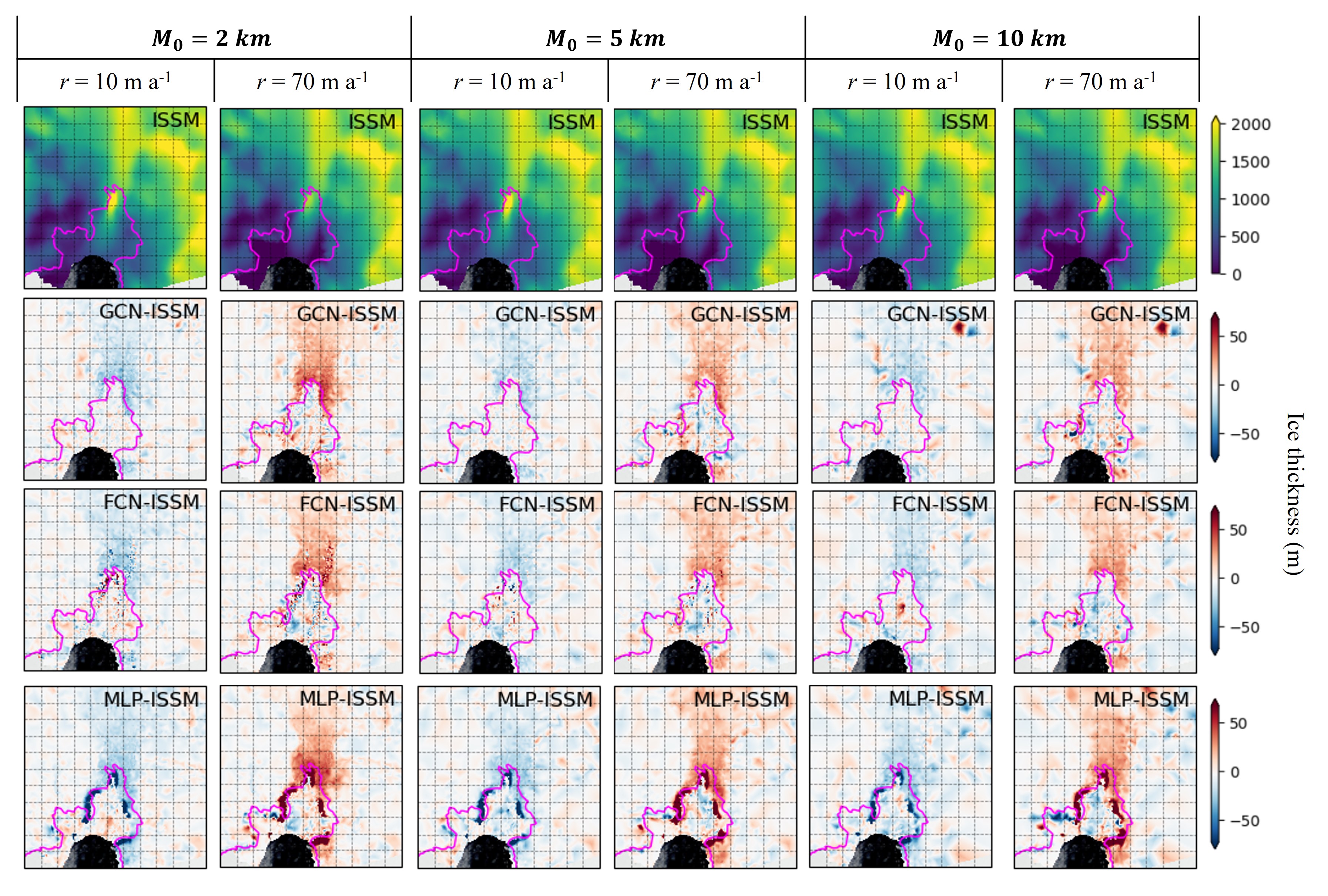} 
    \caption{Maps of ice thickness modeled by the ISSM simulation and difference with deep learning emulators (GCN, FCN, and MLP from top to bottom) for two different basal melting rates ($r$=10 and 70 m $\text{a}^{-1}$) and different initial mesh sizes ($M_0$=2, 5, and 10 km). Each map shows the 20-year average of ice thickness. The ice thickness maps for years 1, 10, and 20 are shown in Figure \ref{A1}. The dashed grids indicate a Cartesian 20 km grid, and the magenta line indicates the grounding line. The south side of the grounding line is floating, and the north side of the grounding line is grounded.}
    \label{Thickness_map}
\end{figure}

The maps of ice thickness from the ISSM simulation and deep learning emulators are depicted in Figure \ref{Thickness_map}. The overall spatial distribution of ice thickness is well reproduced by deep learning emulators: thicker ice ($>$ 1000 m) on the northeast side and thinner ice on the south side near the coast ($<$ 500 m). The emulators tend to overestimate ice thickness at a higher melting rate ($r$ = 70 m $\text{a}^{-1}$) and underestimate ice thickness at a lower melting rate. It is noted that such overestimation and underestimation occur near the central ice stream where the ice moves fast. Additionally, compared to the GCN and FCN, the MLP shows higher errors around the grounding lines. Such a higher error from MLP can be attributed to the no-connectivity characteristics of the MLP architecture, which prohibits the MLP from predicting the complicated interaction between grounding and floating ice around the grounding line. On the other hand, the GCN and FCN reproduce relatively stable accuracy near the grounding line due to their ability to leverage neighboring node information.


\subsection{ Ice velocity}



RMSE and R of ice velocity are shown in Table \ref{table2}. All deep learning models show good agreements with the ISSM simulation results with R $>$ 0.99, but the GCN outperforms the others on all meshes. In particular, the superiority of GCN over FCN and MLP is more evident on a finer mesh of 2 km $M_0$: the RMSE of GCN is lower than the FCN and MLP by 30 m $\text{a}^{-1}$ and 45 m $\text{a}^{-1}$, respectively. Similar to the ice thickness result, the GCN and MLP show relatively low variations in RMSE by different mesh resolutions: the range of RMSE is around 6 m $\text{a}^{-1}$ and 8 m $\text{a}^{-1}$ for GCN and MLP, respectively. However, the RMSE of FCN varies from 58 m $\text{a}^{-1}$ on 10 km $M_0$ to 86 m $\text{a}^{-1}$ on 2 km $M_0$, equivalent to 28 m $\text{a}^{-1}$ of range. As already discussed in the ice thickness result, the fixed 2 km resolution of FCN can lose the detailed ice dynamics at faster ice zones, leading to lower accuracy at finer mesh resolutions.

\begin{table}
\caption{Accuracy of ice velocity for three deep learning emulators with different mesh resolutions. All metrics are averaged for the 10, 30, 50, and 70 m $\text{a}^{-1}$ melting rates. The best accuracy for each mesh size is highlighted in bold.}
\begin{tabular}[t]{c|c|c|c|c|c|c}
\hline 
\multirow{2}{*}{Model} &
\multicolumn{2}{c|}{$M_0$ = 2 km} &
\multicolumn{2}{c|}{$M_0$ = 5 km} &
\multicolumn{2}{c}{$M_0$ = 10 km} \\
\cline{2-7}
&RMSE (m $\text{a}^{-1}$) &R &RMSE (m $\text{a}^{-1}$) &R &RMSE (m $\text{a}^{-1}$) &R\\
\hline
GCN &\textbf{55.668} &\textbf{0.9975} &\textbf{49.607} &\textbf{0.9982} &\textbf{51.341} &\textbf{0.9983} \\
FCN &86.223 &0.9948 &69.982 &0.9969 &58.393 &0.9978\\
MLP &101.187 &0.9920 &94.199 &0.9932 &93.029 &0.9942\\

\hline
\end{tabular}
\label{table2}
\end{table}%



\begin{figure}
    \centering
    \includegraphics[width=0.9\linewidth]{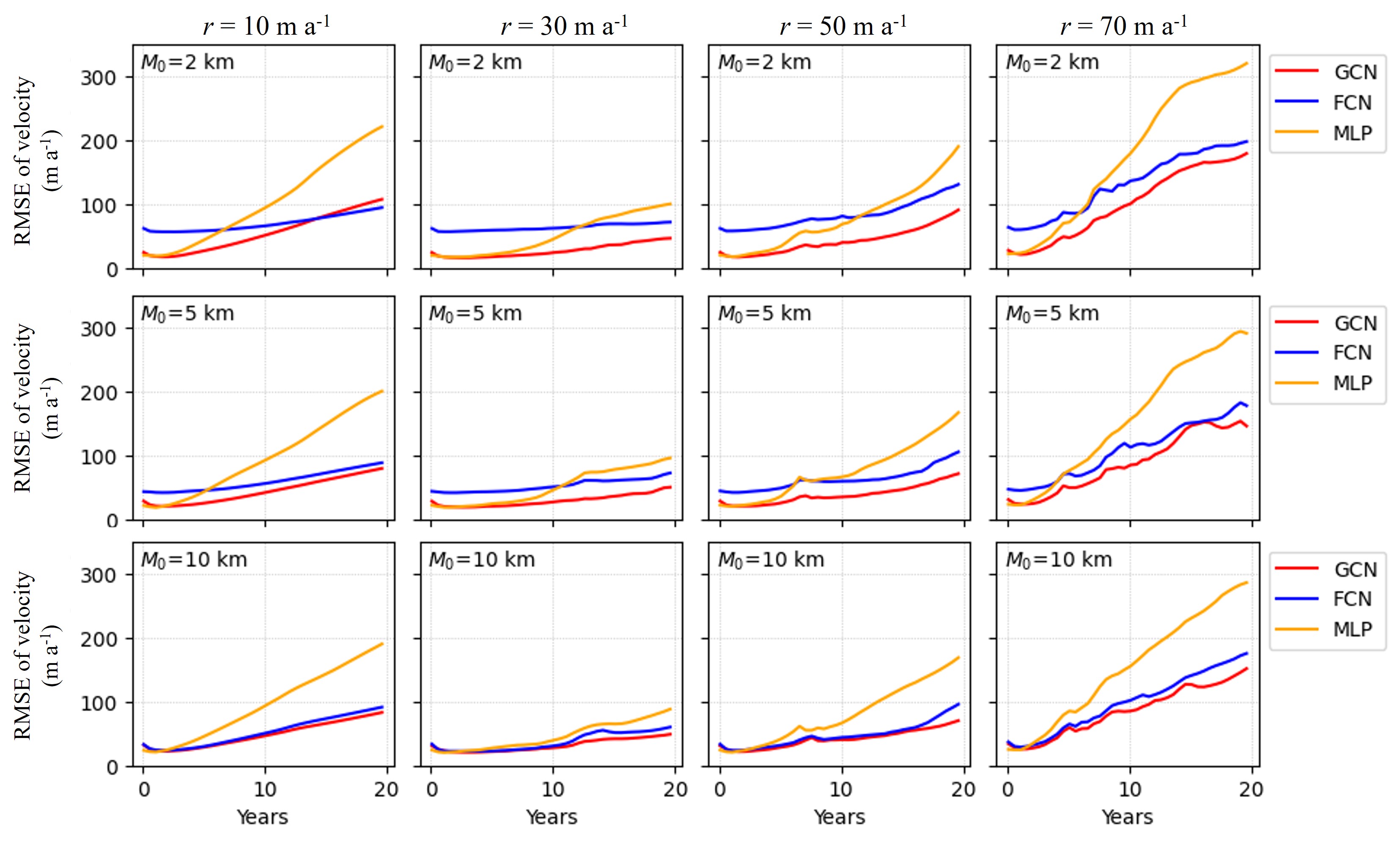} 
    \caption{RMSE of ice velocity by years for different basal melting rates ($r$) and initial mesh sizes ($M_0$).}
    \label{Velocity_year}
\end{figure}

Figure \ref{Velocity_year} shows the RMSE of ice velocity by years for different melting rates. In general, the GCN shows the lowest RMSE over 20-year transient simulation for most of the melting rates and mesh sizes. Similar to the ice thickness result, mesh resolutions do not have a significant impact on the RMSEs of GCN and MLP, whereas the RMSE of FCN decreases with coarse mesh resolutions. In all melting rate scenarios, the RMSEs of emulators increase from the first year to the last year. The MLP shows the most dramatic increase in RMSE by year: at 70 m $\text{a}^{-1}$ of melting rate, the RMSE reaches up to 300 m $\text{a}^{-1}$ in year 20. However, the increases in RMSEs by year for GCN and FCN are not as significant as MLP; at 70 m $\text{a}^{-1}$ of melting rate, the RMSEs reach 150-200 m $\text{a}^{-1}$ in year 20. The higher errors at higher melting rates might be caused by more dynamic ice conditions under more accelerated melting; since ice flows faster as time passes under a higher melting rate, it would be difficult for deep learning emulators to learn such dramatic variations in ice flow.


\begin{figure}
    \centering
    \includegraphics[width=1.0\linewidth]{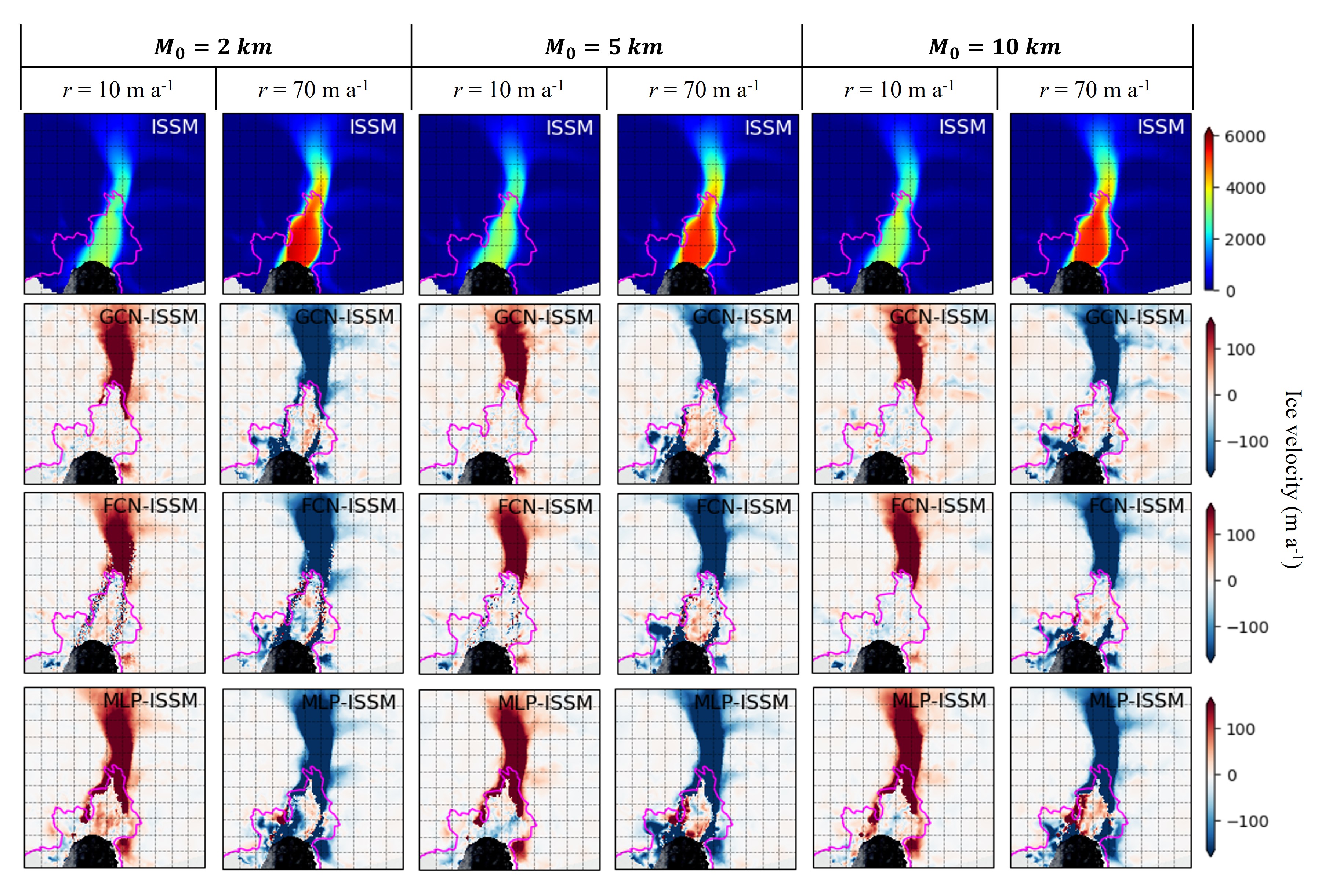} 
    \caption{Maps of ice velocity modeled by the ISSM simulation and difference with deep learning emulators (GCN, FCN, and MLP from up to bottom) for two different basal melting rates ($r$=10 and 70 m $\text{a}^{-1}$) and different initial mesh sizes ($M_0$=2, 5, and 10 km). Each map shows the 20-year average of ice velocity. The ice velocity maps for years 1, 10, and 20 are shown in Figure \ref{A2}. The dashed grids indicate a Cartesian 20 km grid, and the magenta line indicates the grounding line. The south side of the grounding line is floating, and the north side of the grounding line is grounded.}
    \label{Velocity_map}
\end{figure}

The spatial distribution of ice velocity from the ISSM simulation and deep learning emulators are shown in Figure \ref{Velocity_map}. The overall spatial distribution of ice velocity is well reproduced by deep learning emulators: fast ice along the central ice stream, slow ice flow around ice margins, higher velocity under higher melting rates, and increase in ice velocity by time under a higher melting rate. Interestingly, most large errors are found along the grounding part of the central ice stream. All deep learning emulators overestimate velocity at a melting rate of 10 m $\text{a}^{-1}$ and underestimate velocity at a melting rate of 70 m $\text{a}^{-1}$. Similar to the findings from ice thickness results, the MLP exhibits larger errors around the grounding line, which may be caused by the no connectivity propagation process of MLP. Additionally, it is worth mentioning the substantial differences between GCN and FCN results on a mesh resolution of $M_0$ = 2 km. In the $M_0$ = 2 km result, the FCN shows a significant error around the boundary between fast and static ice on the floating ice part, which is not observed in the GCN result. This result implies the limitation of fixed-resolution FCN in describing the detailed ice dynamics with a finer resolution. While the ISSM simulation refines mesh sizes and renders a finer mesh in the fast ice region, the FCN uses a 2 km grid size for all locations regardless of ice dynamic conditions. Hence, the 2 km fixed resolution of FCN does not capture the dynamic details produced by numerical ice sheet models near the boundary of fast and slow ice.


\subsection{ Computational performance}

The main contribution of the deep learning emulators is to reduce the computational time by exclusively using GPUs. The total elapsed times for generating the transient results from the ISSM and deep learning models are shown in Table \ref{table3}. Most deep learning emulators take less time than the ISSM with both CPU and GPU. The MLP generally takes the least time for all mesh sizes because of its simplest architecture. While the computational times of FCN remain consistent for all mesh size experiments because the grid size is always fixed to 2 km, the computational times of GCN and MLP increase with more nodes with a finer mesh resolution. When a GPU is used, the FCN shows approximately 69 times speed-up compared to the ISSM simulations for all mesh resolutions. On the other hand, the GPU-based computation time of MLP is 432 times, 603 times, and 732 times faster than ISSM for 2 km, 5 km, 10 km of $M_0$, respectively; the GPU-based computation time of GCN is 64 times, 86 times, and 103 times faster than ISSM for 2 km, 5 km, 10 km of $M_0$, respectively. When comparing the CPU and GPU computation times for deep learning models, the GPU outperforms the CPU significantly, speeding up by 3-30 times. The most dramatic speed-up by replacing GPU with CPU occurs with FCN: the computational time of FCN speeds up by 28 times using CPU for all mesh resolutions. 

This result demonstrates the computational flexibility of GCN to mesh resolution and the inflexibility of FCN when using different mesh resolution settings. Since the FCN should be implemented on regular grids for all locations with a fixed resolution, it cannot efficiently allocate computational resources in accordance with ice dynamic conditions and computational complexity. However, since the GCN is implemented on adjusted meshes, computational resources can be efficiently allocated to where a more detailed resolution and computational complexity are necessary. In this aspect, we can adjust the mesh resolution of the GCN emulator depending on the desired details of the ice dynamics, whereas we should use the same resolution for all locations in the FCN emulator. Consequently, the GCN would be a more efficient option than the FCN in emulating numerical ice sheet models without losing detailed fine-resolution information.



\begin{table}
\caption{Total computational time (in seconds) for producing final ice sheet transient simulations for 20 years and 36 different melting rates. The fastest emulator is highlighted in bold.}
\begin{tabular}{c|c|c|c|c|c|c|c}
\hline
\multirow{2}{*}{} &
\multirow{2}{*}{Model} &
\multicolumn{2}{c|}{$M_0$ = 2 km}&
\multicolumn{2}{c|}{$M_0$ = 5 km}&
\multicolumn{2}{c}{$M_0$ = 10 km}\\\cline{3-8}
& &CPU &GPU &CPU &GPU &CPU &GPU \\
\hline
Simulation & ISSM &1538.17 & - & 753.30 & - & 541.98 & - \\
\hline
\multirow{3}{*}{Emulator}& GCN &148.19 &23.85 & 69.74 & 17.83 & 45.04 & 14.96 \\
 & FCN &621.72 &22.20 &613.33 &22.54 &612.03 &22.46 \\
 & MLP &\textbf{56.64} &\textbf{3.56} &\textbf{27.45} &\textbf{2.55} &\textbf{18.89} &\textbf{2.10} \\
\hline
\end{tabular}
\label{table3}
\end{table}%

\subsection{ Sensitivity of ice sheet behaviors to basal melting rate}

\begin{figure}
    \centering
    \includegraphics[width=1.0\linewidth]{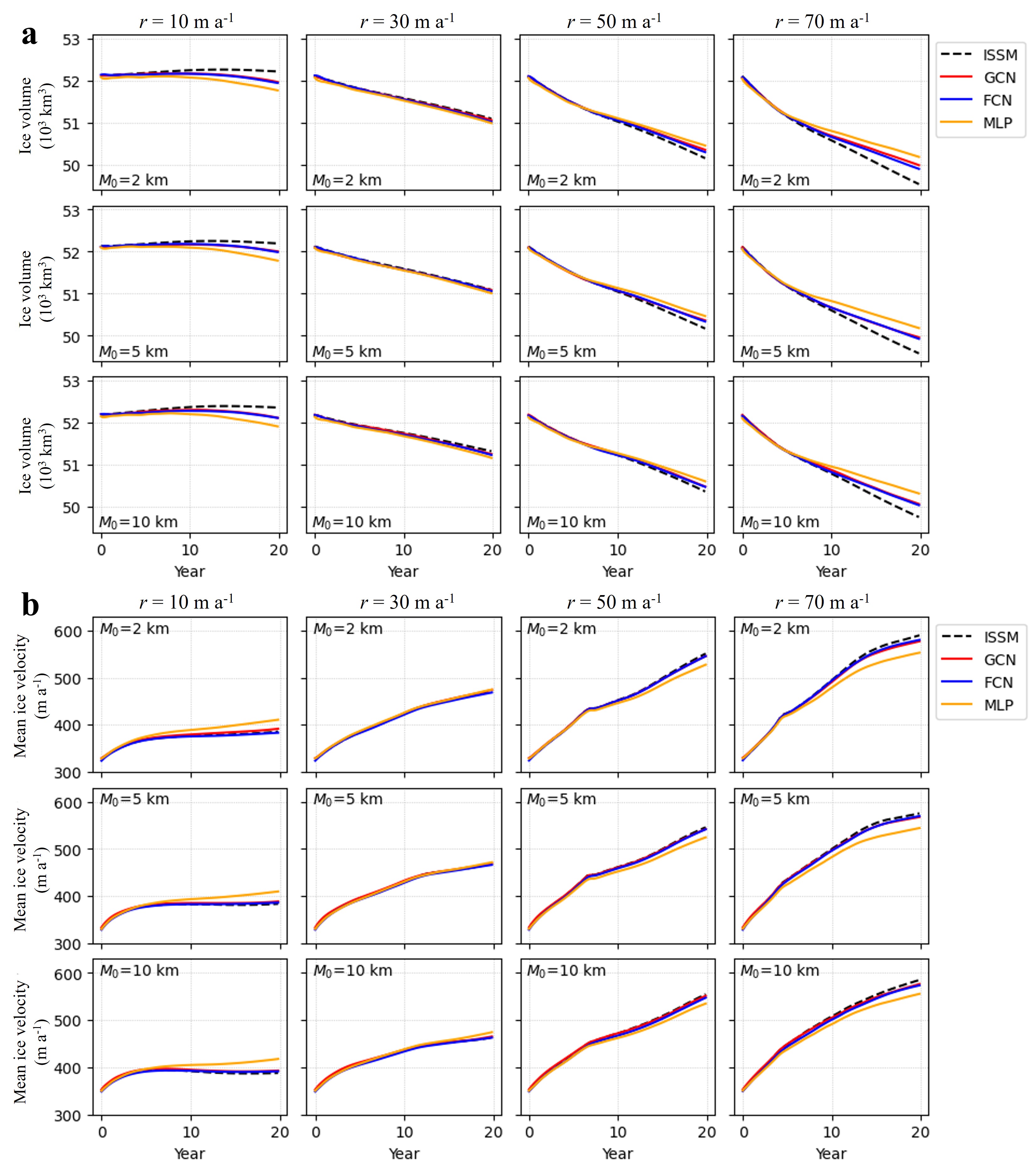} 
    \caption{(a) Total ice volume and (b) mean ice velocity of the PIG over 20 years with four test melting rates ($r$=10, 30, 50, and 70 m $\text{a}^{-1}$) and three initial mesh sizes ($M_0$=2, 5, and 10 km).}
    \label{sensitivity}
\end{figure}

Since the deep learning emulators reduce computational time dramatically, they can be useful for a fast sensitivity analysis to investigate the impacts of environmental parameters on the behavior of ice sheets. Considering that accelerated ocean warming in the Amundsen Sea is the main driver for ice shelf melting in the west Antarctic during the 21st century \citep{Naughten2023, Naughten2018, Jourdain2022}, we use our emulators to examine how the total ice volume and mean ice velocity change with different basal melting rate scenarios. The 20-year variations in total ice volumes and mean ice velocities by different melting rates 10, 30, 50, and 70 m $\text{a}^{-1}$ are retrieved by the ISSM simulations and deep learning emulators (Fig. \ref{sensitivity}). Regarding the total ice volume (Fig. \ref{sensitivity}a), deep learning models successfully reproduce the decrease in ice volume with higher melting rates. As the melting rate changes from 10 to 70 m $\text{a}^{-1}$, the ice volume loss is more accelerated. If the melting rate increases from 10 to 70 m $\text{a}^{-1}$, the PIG will lose approximately 2,600 $\text{km}^3$ of ice after 20 years, according to the ISSM simulations. Such potential ice mass loss changes driven by melting rates are underestimated by deep learning emulators: 2,000 $\text{km}^3$ of ice loss by GCN and FCN, and 1,600 $\text{km}^3$ by MLP. This is because deep learning emulators tend to underestimate the total ice volume at lower melting rate scenarios and overestimate ice volume at higher melting rate scenarios.

We also find the ice velocity increases with higher melting rates (Fig. \ref{sensitivity}b), which are reproduced by the ISSM and all machine learning emulators. If the melting rate changes from 10 to 70 m $\text{a}^{-1}$, the mean ice velocity increases by up to 200 m $\text{a}^{-1}$ after 20 years. In general, deep learning emulators overestimate ice velocity at lower melting rate scenarios and underestimate it at higher melting rate scenarios. Similar to the ice thickness experiment, the MLP shows larger variations of ice velocity with the ISSM simulation, but the GCN and FCN follow the trend of ISSM well with negligibly low differences.


\section{ Discussion}\label{discussion}

\subsection{ Upstream training time}

As shown in Table \ref{table3}, the deep learning emulators can produce the transient simulation results extremely fast by leveraging GPUs. However, since such statistical emulators should be first trained with the numerical simulation data before applying them, it is necessary to consider the upstream training cost to assess whole-process efficiencies. Table \ref{table_training} shows the number of learnable parameters and training time for the GCN, FCN, and MLP models. These models are trained on the Texas Advanced Computing Center (TACC) Lonestar 6 multiple-GPU system equipped with 3 NVIDIA A100 GPUs (40 GB memory). The MLP takes the least training time because of the simplest architecture, followed by GCN and FCN. It is noted that the FCN takes 2.66 times and 4.18 times more training time than GCN and MLP, respectively, because it has more than double the learnable parameters compared to the other emulators. Therefore, considering the model fidelity (Table \ref{table1} and \ref{table2}), the computational time for producing transient results (Table \ref{table3}), and training time (Table \ref{table_training}) together, the FCN would not be the best option to replicate the numerical ice sheet simulations. Instead, depending on the expected level of fidelity and computational speed, the GCN can be a better alternative for representing mesh structures of the numerical ice sheet models.

\begin{table}
\centering
\caption{Upstream computational time for training deep learning models}
\begin{tabular}[t]{c|c|c}
\hline
Model & Number of learnable parameters & Training time (seconds)\\
\hline
GCN & 67,843 & 2324.86\\
FCN & 155,267 & 6195.00\\
MLP & 67,843 & 1482.52\\
\hline
\end{tabular}
\label{table_training}
\end{table}%

\subsection{ Advantages and challenges of GCN}
The first and foremost advantage of the GCN over the traditional FCN or MLP is that the GCN shows better fidelity for the refined mesh structure of ISSM. Since the FCN interpolates the raw ISSM mesh into the regular mesh, the interpolated grid loses the topographical detail and deteriorates the model accuracy for where the ice moves fast. On the other hand, the ice thickness and velocity predicted by the MLP show significant errors with the ISSM results due to the lack of embedded connections between nodes. Hence, the GCN is a better approach to reproducing ice thickness and velocity without losing significant spatial information between nodes. Second, the GCN is flexible to any graph structure and mesh resolution within the area of interest (PIG in this case). Since the FCN uses the fixed regular grid and kernel size, which are not flexible to different resolutions, it is necessary to generate another grid and train another FCN model if we want to change the targeted resolution. Moreover, as FCN resolution increases, the computational demands rise exponentially due to the additional interpolation and excessive number of processed points. On the contrary, once a GCN model is trained, it can be applied to any graph structure of any mesh resolution without a significant variation in model fidelity by mesh resolution (Table \ref{table1} and \ref{table2}). Such flexibility to mesh resolution of GCN allows us to save a lot of computational costs spent for training the model and producing transient simulation results. Finally, although we use the simplest GNN architecture, GCN, in this study, GNNs can be easily adapted and incorporated with different architectures. In general, a GNN architecture is determined by what message-passing procedures are used between nodes (Eq. \ref{eq_gcn_propogation} and \ref{eq_gcn}). Recently, various GNN architectures have been proposed with numerous attention and aggregation algorithms to improve the model for a certain application of interest \citep{Velickovic2018_GAT, hamilton2017_sage, Chebyshev2022}. Hence, based on the GCN architecture proposed in this study, various GNN architectures can be further adapted and developed for ice sheet problems.



However, several challenges of GCN should be considered and mitigated for future applications in ice sheet dynamics. First, like all types of data-driven machine learning approaches, the generalizability is dependent on training samples. Even though our emulators can guarantee substantial fidelity in the PIG region, there is a possibility that these emulators will not perform as well for the other out-of-training ice sheets that have different distributions of ice velocity, ice thickness, and bed topography. Therefore, in order to improve the generalizability of the model, it is necessary to collect training data from various ice sheets and shelves under various environmental conditions. Incorporating physical knowledge (e.g., mass conservation law, SSA equations) into the GNN architecture (so-called physics-informed neural network \citep{He2023, Riel2021, Riel2023, RAISSI2019_PINN}) can also help enhance the generalizability. In addition, since we employ the static graph structure over the 20-year time period, the mass balance changes of the ice sheet caused by the ice sheet domain changes or complex ice dynamical events (e.g., movement of the ice front, calving) cannot be fully described by our emulators. To describe the spatiotemporal changes of ice conditions and boundaries, we can consider using further advanced GNN architectures, such as equivariant GCN (EGCN) \citep{Satorras2021_EGCN}, which preserves the equivariance of graph structures on dynamic systems, or recurrent GNN, which captures the temporal features in graphs \citep{Wu2021}.





\section{ Conclusion}
We propose a high-fidelity and computationally efficient graph convolutional network (GCN) emulator for the Ice-sheet and Sea-level System Model (ISSM), which operates on unstructured meshes. Selecting the Pine Island Glacier (PIG), where ice flows fastest in Antarctica driven by basal melting, as the test site, we train and test a GCN architecture and compare it with traditional non-graph deep learning models, including fully convolutional network (FCN) and multi-layer perceptron (MLP). While the FCN intrinsically uses the fixed kernel convolution for all grid nodes (i.e., fixed connectivity) and the MLP does not use any connectivity information between nodes (i.e., no connectivity), the GCN can leverage the connection between neighboring nodes via the adjacency matrix of graph structure (i.e., flexible connectivity). Based on the advantages of representing adjusted mesh structures of ISSM, the GCN successfully reproduces the 20-year ice thickness and velocity modeling, outperforming the other baseline emulators. Compared to the FCN, which has been commonly used as a deep learning emulator, the robustness of GCN remains consistent regardless of mesh sizes. In particular, whereas the FCN loses fine-resolution ice dynamics in fast ice regions, the GCN can keep the fine-resolution ice dynamics because it directly uses the refined mesh structures of ISSM. The GCN also shows better accuracy in ice thickness and velocity than the MLP, especially near the grounding line, where the dynamic ice behavior changes rapidly by interacting with floating and grounding ice parts. In terms of computational efficiency, the GCN is 60-100 times faster than ISSM simulations by leveraging the parallel processing ability of GPU. This study exhibits for the first time that GCNs have great potential as a statistical emulator that mimics finite-element-based ice sheet modeling. The high fidelity and computational efficiency of the GCN emulator will be useful for predicting the variations in ice dynamics driven by climatological forcings.

\section*{Acknowledgements}
This work is supported by NSF BIGDATA (IIS-1838230, 2308649) and NSF Leadership Class Computing (OAC-2139536) awards.

\bibliography{igsrefs}   
\bibliographystyle{igs}  

\appendix
\section{ Appendix}
\renewcommand\thefigure{\thesection\arabic{figure}}

Figures \ref{A1} and \ref{A2} show the ice thickness and velocity maps, respectively, for years 1, 10, and 20. The overall errors increase from year 1 to year 20, particularly near the fast ice stream regions.

\setcounter{figure}{0}
\begin{figure}
    \centering
    \includegraphics[width=1.0\linewidth]{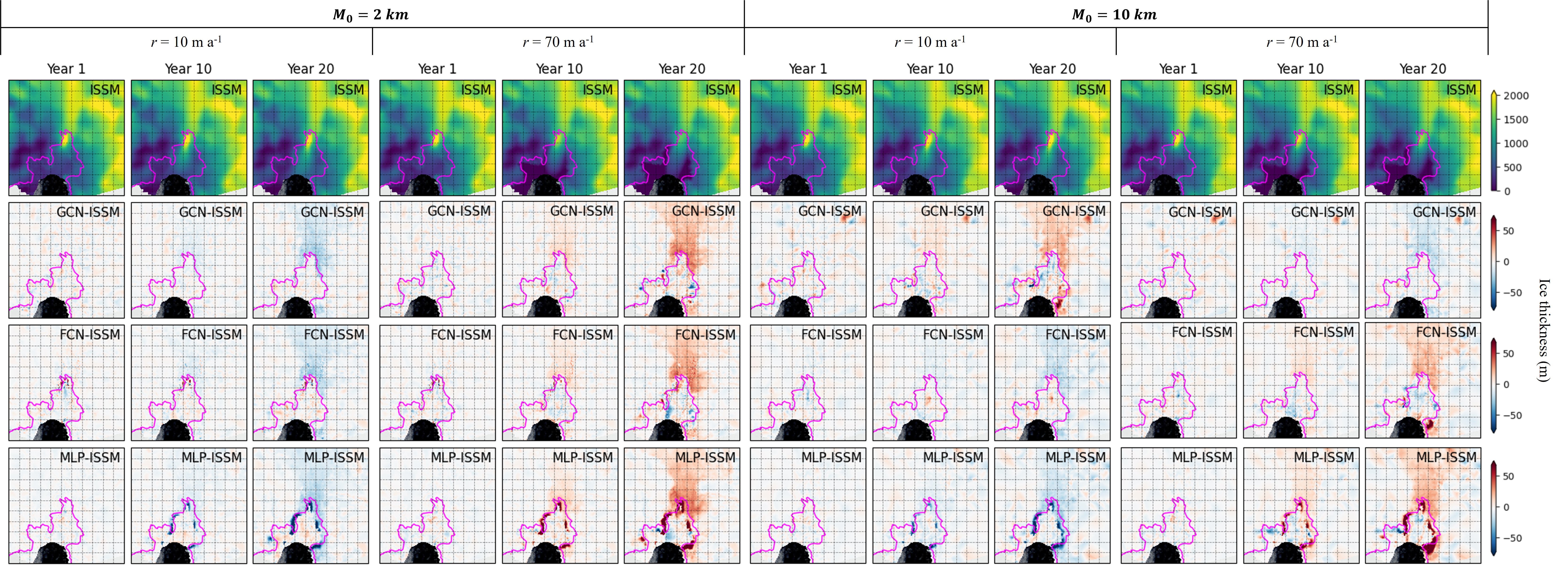} 
    \caption{Maps of ice thickness modeled by the ISSM simulation and difference with deep learning emulators (GCN, FCN, and MLP from up to bottom) for two different basal melting rates ($r$=10 and 70 m $\text{a}^{-1}$) and two initial mesh sizes ($M_0$=2 and 10 km) in years 1, 10, and 20.}
    \label{A1}
\end{figure}

\setcounter{figure}{1}
\begin{figure}
    \centering
    \includegraphics[width=1.0\linewidth]{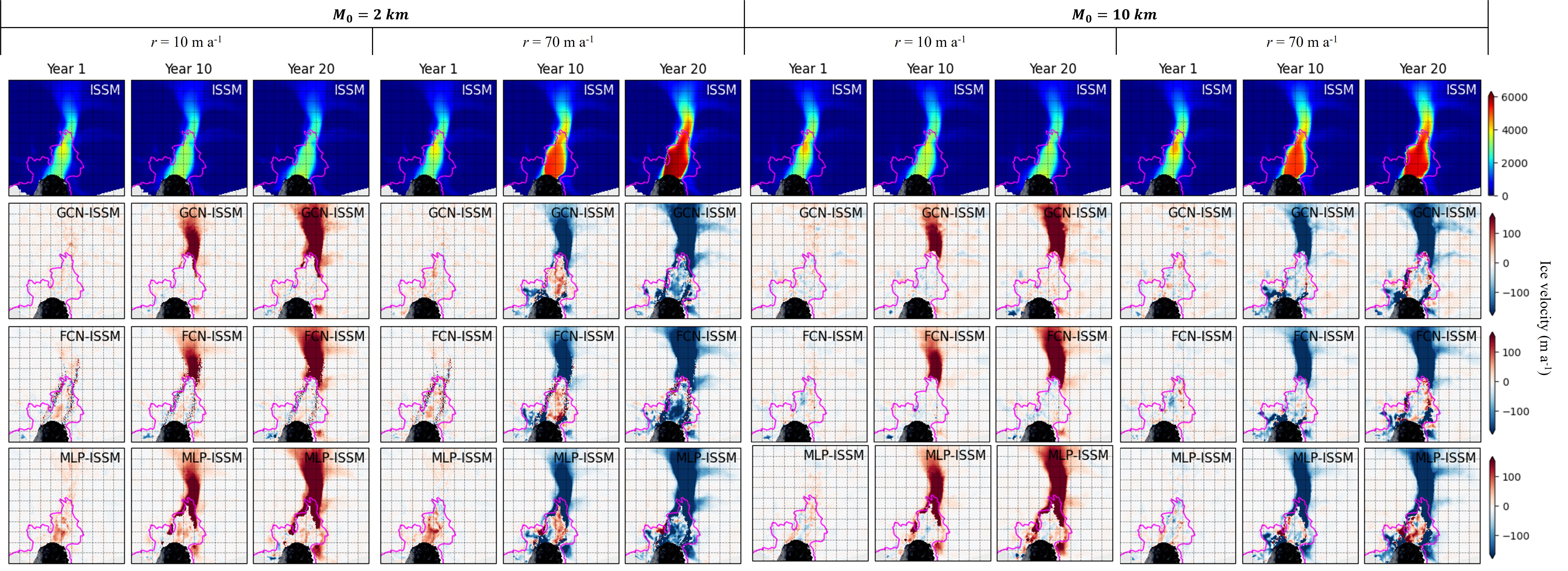} 
    \caption{Maps of ice velocity modeled by the ISSM simulation and difference with deep learning emulators (GCN, FCN, and MLP from up to bottom) for two different basal melting rates ($r$=10 and 70 m $\text{a}^{-1}$) and two initial mesh sizes ($M_0$=2 and 10 km) in years 1, 10, and 20.}
    \label{A2}
\end{figure}

\end{document}